\documentclass[conference]{IEEEtran}

\usepackage{cite}
\usepackage{amsmath,amssymb,amsfonts}
\usepackage{algorithmic}
\usepackage{graphicx}
\usepackage{textcomp}
\usepackage{xcolor}
\usepackage{comment}
\usepackage{todonotes}
\usepackage{wrapfig} \usepackage{booktabs}

\usepackage{svg}
\def\BibTeX{{\rm B\kern-.05em{\sc i\kern-.025em b}\kern-.08em
    T\kern-.1667em\lower.7ex\hbox{E}\kern-.125emX}}

\begin{document}

\title{Scoring with Large Language Models: A Study on Measuring Empathy of Responses in Dialogues}

\author{\IEEEauthorblockN{Henry J. Xie\IEEEauthorrefmark{1}}
\IEEEauthorblockA{\textit{Westview High School} \\
Portland, OR 97229, USA \\
henryjxie@gmail.com}
\and
\IEEEauthorblockN{Jinghan Zhang}
\IEEEauthorblockA{\textit{Portland State University} \\
Portland, OR 97201, USA \\
jinghanz@pdx.edu}
\and
\IEEEauthorblockN{Xinhao Zhang}
\IEEEauthorblockA{\textit{Portland State University} \\
Portland, OR 97201, USA \\
xinhaoz@pdx.edu}
\and
\IEEEauthorblockN{Kunpeng Liu\IEEEauthorrefmark{2}}
\IEEEauthorblockA{\textit{Portland State University} \\
Portland, OR 97201, USA \\
kunpeng@pdx.edu}
}
\maketitle

\renewcommand\thefootnote{}
\footnotetext{* Currently a high school junior; work done as a research intern at PSU}
\footnotetext{\IEEEauthorrefmark{2} Corresponding author}
\renewcommand\thefootnote{\arabic{footnote}}

\begin{abstract}
In recent years, Large Language Models (LLMs) have become increasingly more powerful in their ability to complete complex tasks. One such task in which LLMs are often employed is scoring, i.e., assigning a numerical value from a certain scale to a subject. In this paper, we strive to understand how LLMs score, specifically in the context of empathy scoring. We develop a novel and comprehensive framework for investigating how effective LLMs are at measuring and scoring empathy of responses in dialogues, and what methods can be employed to deepen our understanding of LLM scoring. Our strategy is to approximate the performance of state-of-the-art and fine-tuned LLMs with explicit and explainable features. We train classifiers using various features of dialogues including embeddings, the Motivational Interviewing Treatment Integrity (MITI) Code, a set of explicit subfactors of empathy as proposed by LLMs, and a combination of the MITI Code and the explicit subfactors. Our results show that when only using embeddings, it is possible to achieve performance close to that of generic LLMs, and when utilizing the MITI Code and explicit subfactors scored by an LLM, the trained classifiers can closely match the performance of fine-tuned LLMs. We employ feature selection methods to derive the most crucial features in the process of empathy scoring. Our work provides a new perspective toward understanding LLM empathy scoring and helps the LLM community explore the potential of LLM scoring in social science studies.\footnote{~Code: https://github.com/henryjxie/Scoring-with-Large-Language-Models}
\end{abstract}

\section{Introduction}
\begin{figure}
    \vspace{-6mm}
    \centering
    \includegraphics[width=0.95\linewidth]{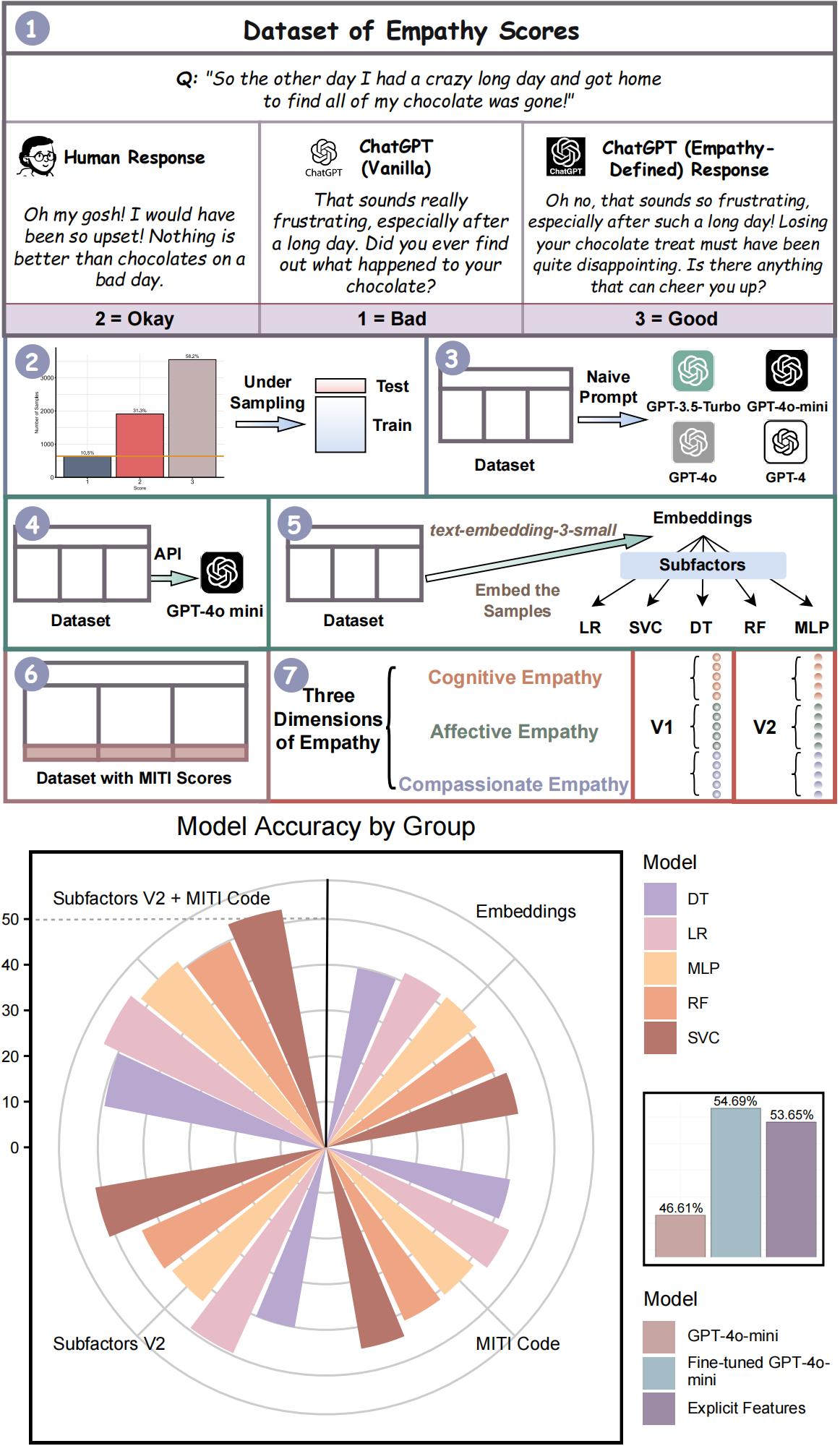}
    \caption{Our Methodology: Dataset, Models, Feature Sets, and Steps (Upper); Scoring Accuracy Achieved with Different Models and Feature Sets (Lower)}
    \label{fig:accuracybygroup}
    \vspace{-6mm}
\end{figure}
In recent years, Large Language Models (LLMs) such as Gemini~\cite{gemini}, GPT-4~\cite{gpt4}, and LLaMA~\cite{llama} have revolutionized Natural Language Processing with their impressive capabilities beyond basic text generation and translation. One such noteworthy capability is scoring, i.e., assigning a numerical value from a certain scale to a subject; specifically, their capability to measure and score empathy, a key aspect of human communication and interaction. Empathy scoring involves understanding and recognizing different dimensions of empathy: cognition, emotion, and compassion. We pose two key research questions: How accurate are LLMs at measuring and scoring empathy as compared to human evaluators and how do we comprehend the specifics involved in LLM scoring?

Empathy describes the understanding of others' thoughts, emotions, experiences, etc. Humans develop their sense of empathy overtime through continual social exposure~\cite{Decety-Lamm-2006}. In contrast, LLMs develop their sense of empathy through recognizing patterns in human conversations and interactions. This may result in situations where LLMs have a divergent understanding and view of empathy. In many areas where LLMs are integrated, empathy is crucial for effective communication. Empathy is used to connect to people on an emotional level, allowing a higher level of understanding. As such, it is important that LLMs have the ability to empathize at a proficient level as they are further integrated into our society.

In this paper, we put LLMs’ comprehension of empathy to the test through its accuracy in empathy scoring  as compared to the gold standard: humans. Empathy scoring involves rating a response to a speaker's utterance based on how empathetic the given response is in the context of the dialogue and the speaker’s utterance. Having LLMs score empathy is a way to understand how LLMs internalize and quantify empathy, especially how they measure it. 

Currently, we have a limited understanding of LLMs' capabilities in empathy scoring. To dive into the behind-the-scenes, we propose a novel and comprehensive framework as shown in Figure~\ref{fig:accuracybygroup}, which pursues the strategy of approximating empathy scoring performance of LLMs and their fine-tuned versions with explicit, explainable, and transparent features of dialogues. We (1) adopt an empathetic dialogue dataset augmented with human empathy scores of responses, (2) balance this dataset with respect to empathy scores, (3) measure the performances of state-of-the-art LLMs, and (4) select the best performing LLM as our base model for investigating empathy scoring. We then introduce different methods at understanding LLM empathy scoring. Our methods include training classifiers (5) with the embeddings of the dialogues and (6) using the Motivational Interviewing Treatment Integrity (MITI) Code~\cite{MITI421} of the dialogues provided in the dataset. We further (7) create a set of explicit subfactors for empathy scoring and combine these subfactors with the MITI Code of the dialogues in training classifiers.  
Evaluation results of our methods (see Figure~\ref{fig:accuracybygroup}) show that when using only embeddings, the classifiers achieve accuracy close to that of generic LLMs, and when utilizing the explicit subfactors as scored by an LLM with the MITI Code and being optimized via feature selection, the classifiers achieve parity with the accuracy of fine-tuned LLMs. Making the subfactors explicit and analyzing their importance in classification can serve as a vehicle to understand how LLMs apprehend empathy, how LLMs deliberate their empathy scores, and how their knowledge of empathy can be used in different scenarios.


\section{Related Work}

LLMs have significant potential in transforming computational social sciences, including their applications in sociology, political science, and psychology~\cite{ziems-etal-2024-large}. One of such applications is interpretable scoring with LLMs. It is critical to understand how LLMs score and how best to use such scores. Human empathy is a central factor in social interaction~\cite{Decety-Lamm-2006,wang2023emotional}. Measures of empathy, which are central to understanding online conversation, are of particular importance in the growing cyberspace~\cite{POWELL2017137}. There have also been many studies on the generation of empathetic responses in dialogues~\cite{welivita-pu-2020-taxonomy,welivita-etal-2023-empathetic}. Empathetic and emotional paraphrasing have played a key role in empathetic response generations~\cite{Seehausen2012,DBLP:conf/wassa/XieA23}. Empathy measures are critical in understanding how effective these approaches are. Human ratings are the gold standard of scoring empathy~\cite{welivita2024chatgptempathetichumans}, however not scalable. Therefore, automatic empathy measures through interpretable LLM scoring are a highly desirable alternative.  

\section{Problem Description and Dataset} 

\subsection{Problem Description}

\textit{The Task of Empathy Scoring.} The general task of scoring with LLMs involves taking an input text and assigning a numerical score to the input text. Specifically, given a dialogue context $c_i$, a speaker utterance $u_i$, and a response $r_i$ to $u_i$ in the context $c_i$, the task of empathy scoring is to assign a numerical score $s_i$ of how empathetic the response $r_i$ is. The score ranges from 1: bad empathetic response, 2: okay response, and 3: good response. The dialog context $c_i$ can be a description of the conversation situation, the previous utterance of the dialogues, or both. Therefore, a model $M$ for empathy scoring works as follows: $M(c_i, u_i, r_i) \rightarrow s_i, s_i\in \{1, 2, 3\}$. 

\subsection{Dialogue Dataset}
The dataset used in our study is a dataset created by Welivita and Pu~\cite{welivita2024chatgptempathetichumans}. It is sourced from the EmpatheticDialogues dataset~\cite{rashkin-etal-2019-towards} by randomly selecting 2000 dialogues, each with a unique (situation, speaker utterance) pairing. This dataset contains three different responses associated with each (situation, speaker utterance) pair, one from a human, one from ChatGPT, and one from ChatGPT with an empathy-defining prompt. In addition, the dataset contains human ratings of empathy for all the responses. The scores are either a 1, 2, or 3, representing a bad, okay, or good empathetic response respectively. In this study, we utilize the 2000 situation-speaker utterance pairs, the three responses associated with the (situation, speaker utterance) pair, and the human rated empathy scores. Additionally, we utilize the MITI Code provided for each response. The MITI Code is used to determine the effectiveness of a response in the context of mental health. 

\section{Benchmarking LLM Empathy Scoring} 

In this section, we benchmark the empathy scoring capabilities of state-of-the-art LLMs. We first create a unified balanced training and test dataset for consistent evaluation. We evaluate several LLMs to select the best-performing model, and use this model to obtain a baseline accuracy of empathy scoring and then a peak accuracy by fine-tuning this model. We regard this peak accuracy as the gold standard which represents the upper bound of LLM scoring capability with the given dataset.

\subsection{Creating Consistent Training and Test Datasets}
\label{SubSec:Dataset}
For optimal evaluation, it is crucial to have consistent training and test datasets. 
The original dataset contains 6000 unique (situation, speaker utterance, response) triplets. 
There are human-rated empathy scores associated with each triplet. However, this dataset does not contain a balanced number of each score. More than half of the triplets---3549---are associated with a score of $3$, while there are only 1811 dialogues associated with a score of $2$, and 640 dialogues associated with a score of 1. To create a balanced dataset, we undersample the original dataset to contain the same number, i.e., $640$, of 1s, 2s, 3s scores and then split it $80\% / 20\%$ into the training and test datasets. 

\subsection{Selecting the Best Performing LLM Model}

\begin{wraptable}{r}{0.475\columnwidth}
\vspace{-4mm}
\caption{Accuracy on different LLMs}
\centering
\small 
\begin{tabular}{cr}
\toprule
{\bf Model} & {\bf Accuracy}  \\
\midrule
gpt-3.5-turbo & {\bf 0.3854} \\
gpt-4 & 0.4375 \\
gpt-4o & 0.3880 \\
gpt-4o-mini & {\bf 0.4661} \\
\bottomrule
\end{tabular}
\vspace{-2mm}
\label{tab:models}
\end{wraptable}

To determine which LLM is most effective at the empathy scoring task, we conducted an experiment with four popular LLMs from OpenAI shown in Table~\ref{tab:models}. This experiment consists of prompting the four LLMs with instructions on how to score the response, given the situation, speaker utterance, and response. The experiment was done on the unified test dataset created above in Section~\ref{SubSec:Dataset}. 

Table~\ref{tab:models} shows the accuracy of the four LLMs' scores, as compared to the gold standard of human scores. It was evident that the LLMs GPT-3.5-turbo and GPT-4o were the least accurate at scoring empathy, with accuracy of 38.54\% and 38.80\% respectively. The second most accurate model was GPT-4 at 43.75\% accuracy. The most accurate model was GPT-4o-mini, which scored the correct empathy score 46.61\% of the time. GPT-4o-mini is also the cheapest model of the four. Therefore, we chose GPT-4o-mini as our base model.


\subsection{Obtaining a Baseline Accuracy with Naive Prompt}

In the model evaluations above, we utilized what we call the ``Naive'' prompt as shown in the box below. It only contains the basic instruction: assign an empathy score to the response on a scale of 1 to 3 (bad, okay, or good), given the situation, speaker utterance, and response. We use the accuracy of GPT-4o-mini when prompted with the Naive prompt, $46.61\%$, as the baseline accuracy. It serves as the control of our experiments. It also reflects the current capability of LLM empathy scoring when compared to human scoring, showing that the state-of-the-art LLMs are significantly more accurate than random guessing ($33.33\%$) but have considerable room for improvement. 

\begin{center}
    \fbox{\rule[-.8cm]{0cm}{0cm}
    \begin{minipage}[t]{3.2in}
       {\small
        \noindent{\bf Naive Prompt for Empathy Scoring:}  
        You are given a situation context, a speaker utterance, and a response to the speaker utterance in the situation context. Please score the response on a scale of 1 to 3, where a score of 1 means a bad empathetic response, a score of 2 means an okay empathetic response, and a score of 3 means a good empathetic response.
        }
    \end{minipage}
}
\end{center}

\subsection{Obtaining a Peak Accuracy through Fine-tuning} 

For our study, it is important to know the peak accuracy that LLMs can achieve at empathy scoring compared to human scores. In general, fine-tuning an LLM is the way to achieve the best performance in the given task. Fine-tuning allows the LLM, in our study GPT-4o-mini, to be further trained on the empathy scoring dataset. Using \texttt{OpenAI's API}~\footnote{https://platform.openai.com/docs/overview
}, we fine-tuned GPT-4o-mini using the undersampled unified training dataset. We systematically explored different combinations of the hyperparameters \textit{n\_epochs} and \textit{learning\_rate\_multiplier} (LRM) to boost the accuracy of the fine-tuned model. 

\begin{wraptable}{r}{0.475\columnwidth}
\vspace{-5mm}
\caption{Scoring Accuracy of Fine-Tuned GPT-4o-mini Utilizing Different Hyperparameters}
\centering
\small 
\begin{tabular}{crr}
\toprule
{\bf Epoch} & {\bf LRM} & {\bf Accuracy}  \\
\midrule
3 & 1.80 & 0.4844 \\
4 & 0.25 & 0.5417 \\
4 & 0.50 & {\bf 0.5469} \\
4 & 0.75 & 0.5365 \\
4 & 1.00 & 0.5130 \\
\bottomrule
\end{tabular}
\label{tab:fine-tune-models}
\end{wraptable}
The different combinations of epochs and LRMs with the corresponding accuracy of the fine-tuned models can be found in Table \ref{tab:fine-tune-models}. The highest accuracy that we achieved with a fine-tuned model was $54.69\%$, where n\_epochs=4 and the learning\_rate\_multiplier=0.5. This accuracy is roughly $20\%$ higher than arbitrarily guessing ($33.33\%$) and $8\%$ higher than the default accuracy ($46.61\%$) of GPT-4o-mini. This fine-tuned GPT-4o-mini accuracy represents the peak accuracy that the state-of-the-art LLMs can achieve on this given dataset. This result shows that the state-of-the-art LLMs such as GPT-4o-mini, have a strong understanding of how to measure empathy after fine-tuning. 

\section{Understanding LLM Empathy Scoring}

Although LLMs and their fine-tuned versions have demonstrated promising capabilities in empathy scoring, the inner workings of their scoring mechanisms are opaque. It is highly desirable to make LLM empathy scoring more explainable and transparent. Therefore, we propose and explore several methods that can provide more details and perspectives for understanding the inner workings of LLM empathy scoring mechanisms. These methods include training classifiers using embeddings, the MITI Code, explicit subfactors of empathy, and a combination of these methods. 


\subsection{Embeddings} 

\begin{wrapfigure}{r}{0.5\columnwidth}
    \centering
    \includegraphics[width=0.5\columnwidth]{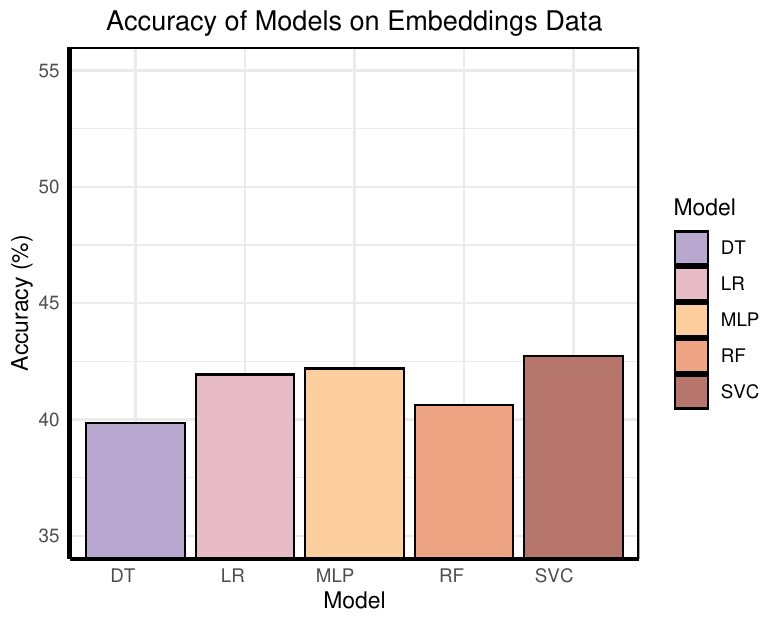}
    \caption{Scoring Accuracy of Classifiers on Embeddings}
     \vspace{-4mm}
    \label{fig:embeddings}
\end{wrapfigure}

We first obtain the embeddings for both the training and test datasets, utilizing the \emph{text-3-embedding-small} embedding model of OpenAI~\cite{openai2024chatgpt}. We then use the obtained embeddings to train five different classifier models: Logistic Regression (LR), Support Vectors (SVC), Decision Tree (DT), Random Forest (RF), and Multilayer Perceptron (MLP)~\cite{scikit-learn}. Using embeddings is an ideal place to start to understand how LLMs score, as the embedding method creates “subfactors” from the text. As shown in Figure \ref{fig:embeddings}, the highest accuracy from embeddings was 42.71\% by SVC, with LR and MLP also having promising accuracy. Considering that OpenAI's embedding API does not employ GPT models, it is reasonable that the result is around $4\%$ lower than the baseline accuracy of GPT-4o-mini. We can view each dimension of the embeddings as a feature. The experimental results reveal that using features may be a promising approach to understanding LLM empathy scoring.



\subsection{Motivational Interviewing Treatment Integrity (MITI) Code}

\begin{wrapfigure}{r}{0.5\columnwidth}
    \vspace{-3mm}
    \centering
    \includegraphics[width=0.48\columnwidth]{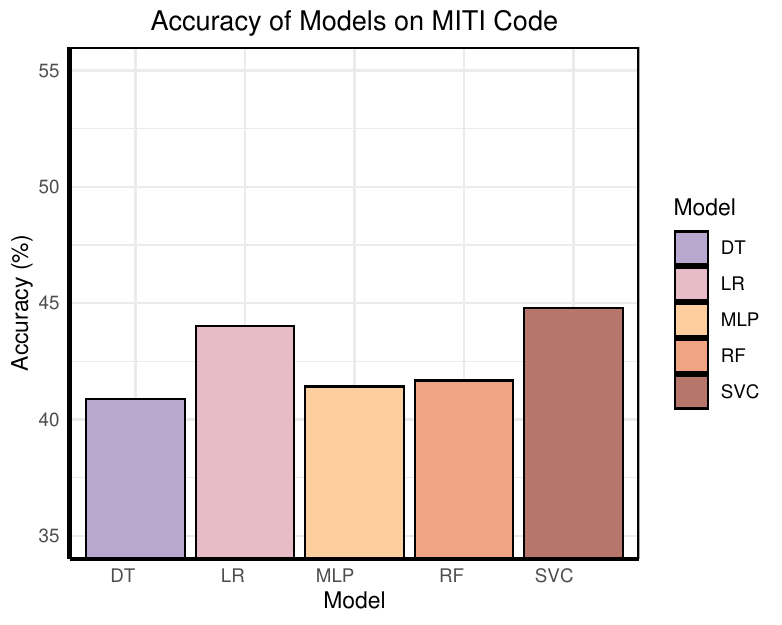}
    \caption{Scoring Accuracy of Classifiers on MITI Code}
    \label{fig:miti}
    
\end{wrapfigure}
Although embeddings are a great starting point, the “features” embedded are uninterpretable to humans. Our next step to further understand LLM empathy scoring is to utilize the MITI Code provided for each (situation, speaker utterance, response) triplet in Welivita and Pu's dataset. The MITI Code is used to evaluate the effectiveness of a mental health professional’s response to a patient. There are 15 different possible codes that a response could be assigned. The dataset provides the codes assigned to each of the responses.
We train classifiers to determine how well the MITI Code represents the responses in terms of empathy scoring. We also evaluate how the classifiers trained using the MITI Code performed as compared to the GPT-4o-mini. As seen in Figure \ref{fig:miti}, the trained classifier with the best performance is SVC at 44.79\% accuracy. 
When compared to the LLM baseline accuracy, they are approximately the same. This illustrates that the MITI Code assigned to the responses do contain information strongly relevant to empathy scoring.

\subsection{Explicit Subfactors}

\begin{figure}[h]
    \vspace{-2mm}
    \centering
    \includegraphics[width=0.98\linewidth]{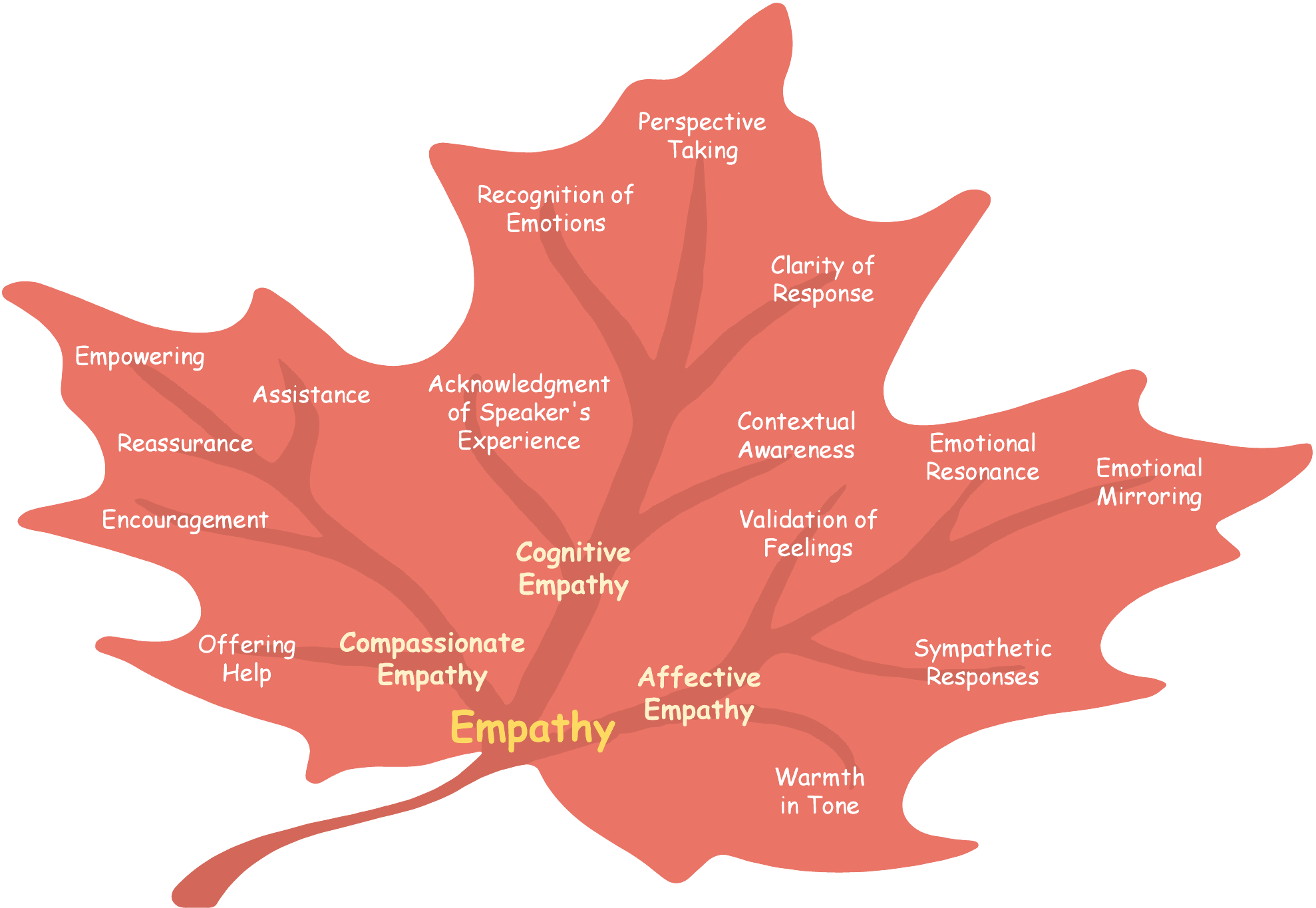}
    \caption{3 Dimensions of Empathy and Their Subfactors}
    \vspace{-8pt}
    \label{fig:3dimssubfactors}
\end{figure}

In this study, we introduce a novel method that utilizes more interpretable subfactors to humans. These subfactors can help better understand how LLMs make decisions on scoring empathy. We use the three dimensions of empathy---Cognitive Empathy, Affective (Emotional) Empathy, and Compassionate Empathy~\cite{Decety-Lamm-2006}---as the basis to build our subfactors. We then generate two different sets of 15 subfactors, five subfactors for each dimension of empathy. The first set of 15 subfactors V1 is generated through prompting ChatGPT-4o to recommend 5 subfactors for each dimension of empathy only once. We then create a second set of ``reinforced'' 
15 subfactors V2, a more refined and robust set of subfactors as shown in Fig.~\ref{fig:3dimssubfactors} where the main stems represent the three empathy dimensions while the branch stems represent the subfactors under each dimension. 
This second set of subfactors is generated by prompting ChatGPT-4o ten times given the dataset and asking it to recommend subfactors based on the dialogue triplets and human-rated scores. We then select the most recurring five subfactors for each dimension of empathy. 

\begin{wrapfigure}{r}{0.5\columnwidth}
    \centering
    \includegraphics[width=0.48\columnwidth]{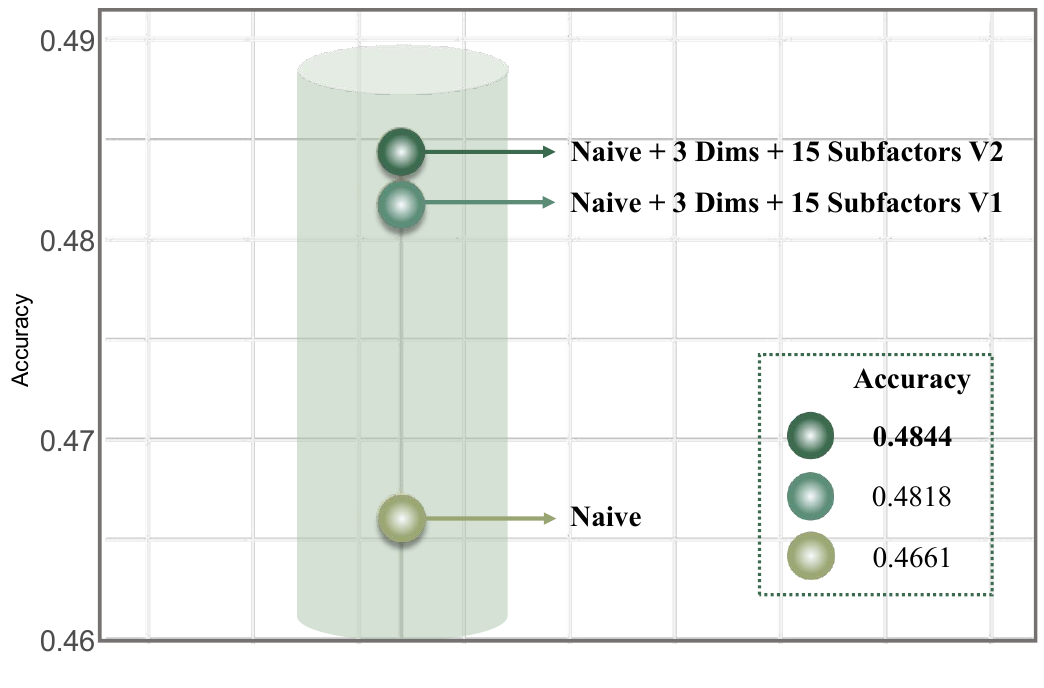}
    \caption{Scoring Accuracy of Different Prompt Combinations on GPT-4o-mini}
    \label{fig:scoringaccuracy4omini}
\end{wrapfigure}

To determine if these subfactors are effective in empathy scoring, we first enhance our Naive prompt for empathy scoring with these 15 subfactors and their definitions provided by ChatGPT. We then re-score the test dataset with the enhanced prompt. As shown in Fig.~\ref{fig:scoringaccuracy4omini}, both versions of the 15 subfactors improve the accuracy of empathy scoring with V2 saw the bigger improvement, so we decided to adopt V2.

\begin{wrapfigure}{r}{0.5\columnwidth}
    \centering
    \includegraphics[width=0.48\columnwidth]{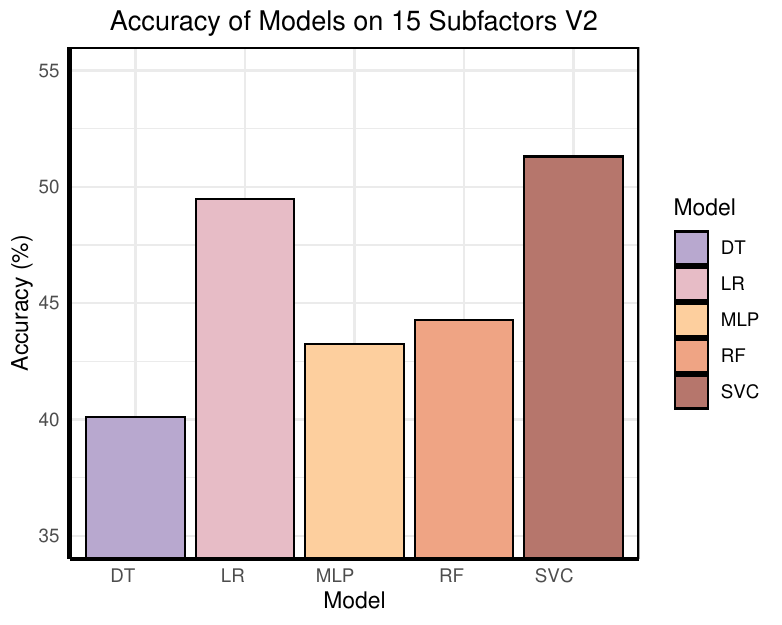}
    \caption{Scoring Accuracy of Classifiers on 15 Subfactors V2}
    \label{fig:subfactors}
\end{wrapfigure}

We then score our dataset with GPT-4o-mini on the 15 Subfactors V2. GPT-4o-mini was prompted to assign a score (ranging from 1 to 10) to each of the 15 subfactors, with a score of 1 representing that the subfactor is not found in the response and a score of 10 being the subfactor is extremely prevalent in the dialogue. With these scores, we then train the classifier models as shown in Figure~\ref{fig:subfactors}. This method allows for a better understanding about GPT-4o-mini’s reasoning when scoring, as we can see how it assigns scores to each subfactor and their influences on the final empathy score. This method yielded a peak accuracy of 51.30\% through the SVC Classifier. 

\subsection{Combining MITI Code and Explicit Subfactors} 

As the above studies show, both the MITI Code and explicit subfactors can achieve satisfactory performance close to the peak performance of GPT-4o-mini. Additionally, both of them can provide explainable and human-interpretable information for empathy scoring. Naturally, it is promising to investigate the possibility of integrating the MITI Code and the 15 explicit subfactors for better model training. 

\begin{wrapfigure}{r}{0.5\columnwidth}
    \centering
    \includegraphics[width=0.5\columnwidth]{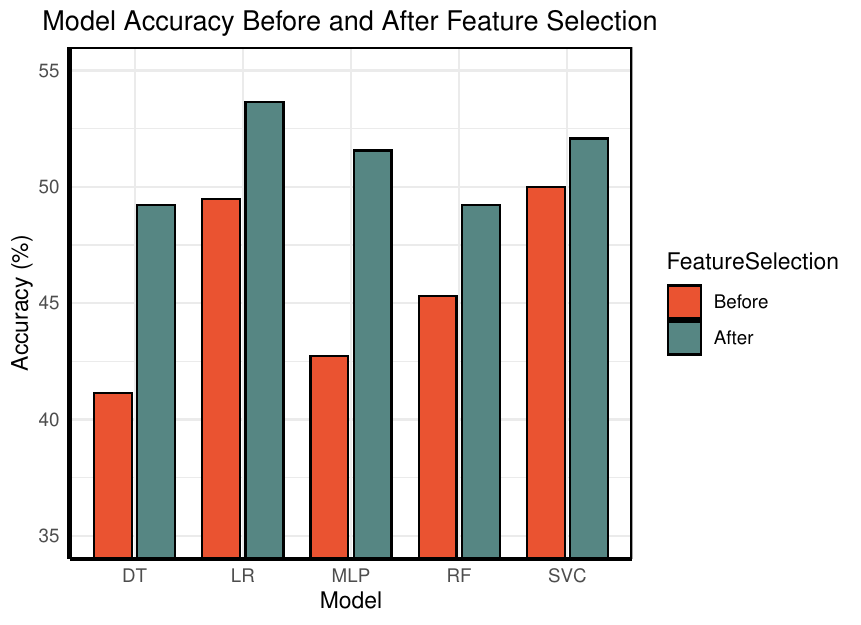}
    \vspace{-3mm}\caption{Accuracy of Regression Models on 15 Subfactors V2 + MITI Code}
    \label{fig:subfactors-miti}
\end{wrapfigure}

We concatenate the MITI Code and the 15 subfactors V2 into a vector of 30 features, and train classifiers using this expanded training data. The results can be found in Figure \ref{fig:subfactors-miti} as noted by the ``before'' label. The highest accuracy of these classifiers was about 50\%, achieved by both LR and SVC, which is worse than expected. The potential reason could be that there exists redundancy in the concatenated vector. Consequently, we try feature selection to eliminate redundancy and select informative features.

\begin{figure}[htb]
\centerline{\includegraphics[width=\columnwidth]{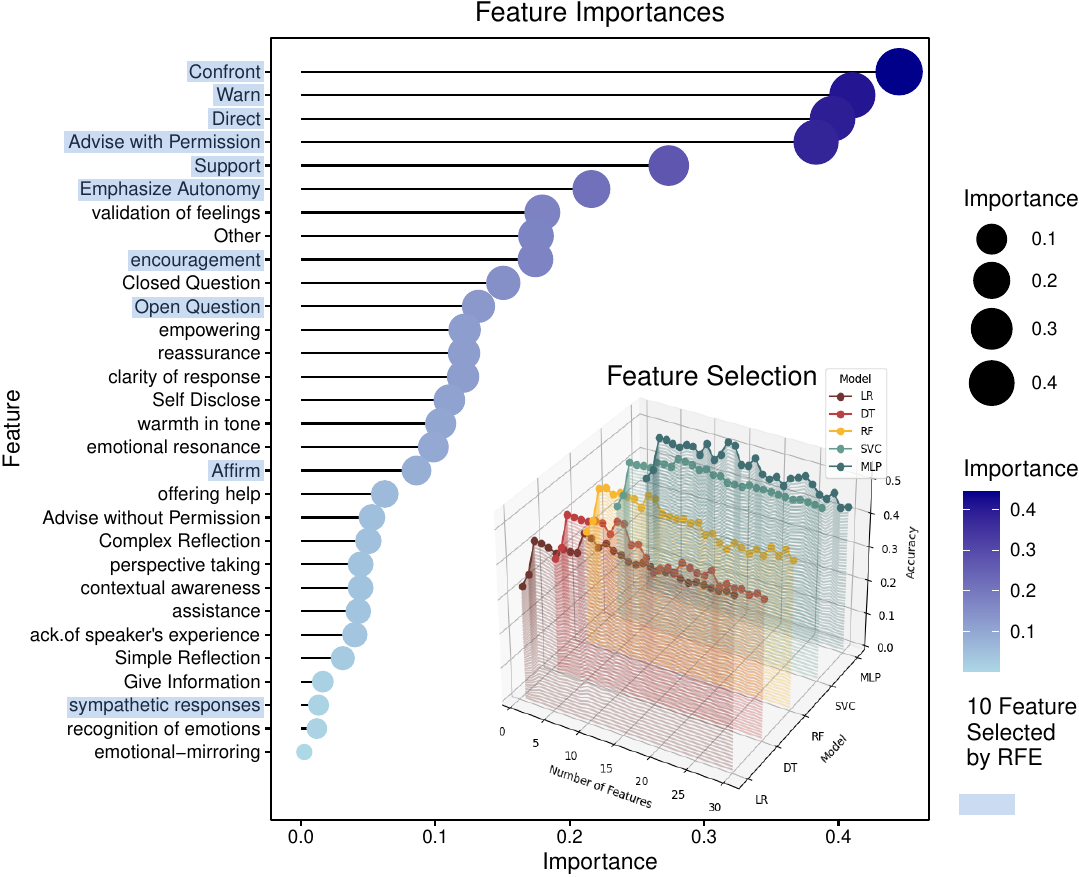}}
\caption{Importance of Different Features}
\label{fig:feature-importance}
\vspace{-10pt}
\end{figure}

We employ the Recursive Feature Elimination (RFE) algorithm for feature selection~\cite{scikit-learn}. The feature space consists of 15 MITI features and 15 explicit subfactors. The selected feature number ranges from 1 to 30. Figure~\ref{fig:feature-importance} (lower right) shows the accuracy achieved by each model when a different number of features were selected. It can be observed that the peak accuracy for each model is achieved when about 10 features are selected. As shown in Figure~\ref{fig:subfactors-miti}, the peak accuracy of each model achieved with RFE improves notably over the base accuracy before feature selection. The top accuracy of 53.65\% is achieved by LR when 10 features are selected. The importance of each feature, which measures how much a feature contributes to classification, can be found in Figure~\ref{fig:feature-importance}, where the capitalized features are part of the MITI Code while the lower case ones are part of 15 explicit subfactors. The selected features (as noted with blue background) are not necessarily the 10 features of the highest importance due to potential duplication and correlation. The results from RFE show that when selecting an optimal set of features from the MITI Code and 15 explicit subfactors for classifier training, the performance of the classifiers can closely approximate the peak performance of the fine-tuned GPT-4o-mini model, and there is a notable increase in accuracy when compared to the classifiers trained without feature selection. Instead of GPT-4o-mini being a black box for empathy scoring, we can now employ the trained classifiers as semi-transparent boxes that lay out what and how features are utilized in scoring. 

\section{Conclusions and Future Work}

In this paper, we develop a framework for understanding LLM empathy scoring through empathy-relevant features. Features considered include embeddings, the MITI Code, the three dimensions of empathy, and a set of 15 explicit subfactors of empathy. Using the 15 explicit subfactors plus the MITI Code, the trained classifiers can effectively reach the peak empathy scoring performance for the state-of-the-art LLMs, even after fine-tuning. This way, our more explicit empathy scoring approach can be utilized in place of direct scoring by LLMs to make empathy scoring more explicit and transparent. 


The dataset utilized in our study contains additional information such as the sentiment and emotion of a dialogue (situation context, speaker utterance, and response). We plan to explore how sentiment and emotion can be factored in empathy scoring. A limitation of this study is that the dialogues are from a single dataset. In future work, we will use dialogues from a variety of data sources to achieve a more representative understanding of empathy scoring that factors in a wide range of backgrounds and contexts.

\bibliographystyle{abbrv}
\bibliography{references.bib} 

\end{document}